\documentclass[sigconf]{acmart}

\usepackage{booktabs} 
\usepackage{multirow}

\begin{document}
\title{Seek and You Will Find: A New Optimized Framework for Efficient Detection of Pedestrian}

\author{Sudip Das}
\orcid{0000-0001-6069-0240}
\affiliation{%
  \institution{Indian Statistical Institute}
  \city{Kolkata}
  \country{India}
  \postcode{}
}
\email{d.sudip47@gmail.com}

\author{Partha Sarathi Mukherjee}
\orcid{0000-0002-9254-5416}
\affiliation{%
  \institution{Indian Statistical Institute}
  \city{Kolkata}
  \country{India}
}
\email{pathosarothimukherjee@gmail.com}

\author{Ujjwal Bhattacharya}
\affiliation{%
  \institution{Indian Statistical Institute}
  \city{Kolkata}
  \country{India}
}
\email{ujjwal@isical.ac.in}


\renewcommand{\shortauthors}{Das et al.}

\begin{abstract}
Studies of object detection and localization, particularly pedestrian detection have received considerable attention in recent times due to its several prospective applications such as surveillance, driving assistance, autonomous cars, etc. Also, a significant trend of latest research studies in related problem areas is the use of sophisticated Deep Learning based approaches to improve the benchmark performance on various standard datasets. A trade-off between the speed (number of video frames processed per second) and detection accuracy has often been reported in the existing literature. In this article, we present a new but simple deep learning based strategy for pedestrian detection that improves this trade-off. Since training of similar models using publicly available sample datasets failed to improve the detection performance to some significant extent, particularly for the instances of pedestrians of smaller sizes, we have developed a new sample dataset consisting of more than 80K annotated pedestrian figures in videos recorded under varying traffic conditions. Performance of the proposed model on the test samples of the new dataset and two other existing datasets, namely Caltech Pedestrian Dataset (CPD) and CityPerson Dataset (CD) have been obtained. Our proposed system shows nearly 16\% improvement over the existing state-of-the-art result.
\end{abstract}

%
%





\maketitle

\section{Introduction} \label{sec:intro}
In the area of computer vision and related research, pedestrian detection is an important object localization problem due to its notable application potentials. Acceptably accurate detection of  pedestrians in video frames of road scenes still remains a challenging problem due to the enormous variations in the instances of pedestrians with respect to their size, pose, lighting condition, proportion of the occluded parts etc. Study of recent literature shows that deep learning based models are capable of producing improved results over the traditional methods. However, there are a few important concerns related to deep learning based strategies. One of them is the complexity of the model in terms of the number of its trainable parameters and consequently requirement of the computational resources. Often such a model requires multiple processing units, considerably large amount of memory and an efficient parallel computation framework for its simulation. Also, a large amount of labelled data is crucial for effective training of the network. Another issue is the trade-off between computational speed measured by FPS (frames processed per second) and the accuracy of estimation of bounding box around the pedestrian figures which is expressed as Miss Rate (MR). It is clear that simultaneous improvement of both FPS and MR is a challenging task. Contribution of the present study is threefold. Firstly, it describes a new deep architecture (detailed in Sec. \ref{sec:proposedmodel}) requiring less computational resources compared to the existing state-of-the-art models. Secondly, our model is capable of optimizing the trade-off between FPS and MR providing  significant improvement in accuracy over existing heavy weight models. This has been described in details in Sec. \ref{sec:experimentalresult}. Finally, we have developed a new sample dataset consisting of little more than 80K annotated pedestrian figures. The images in this dataset are high resolution video frames captured under various traffic conditions. Further details of this new dataset are provided in Sec. \ref{sec:datasetpreparation}.

A state of the art Pedestrian Detection system should have the following qualities:


\begin{itemize}
    \item \textbf{Precision of recognition} in each frame of a video must be guaranteed. This is measured by the well known metric \textbf{Miss Rate}.  In real life scenarios precision of recognition is affected due to unknown number of pedestrians, different size of pedestrian images due to physical dimension and their varying distances from the focus of the camera. Also, recognition accuracy plummets due to poor lighting condition and proportion of occluded part (Fig. \ref{fig:challenges}).
  \begin{figure}[h]
    \begin{center}
          \includegraphics[width=1.5in]{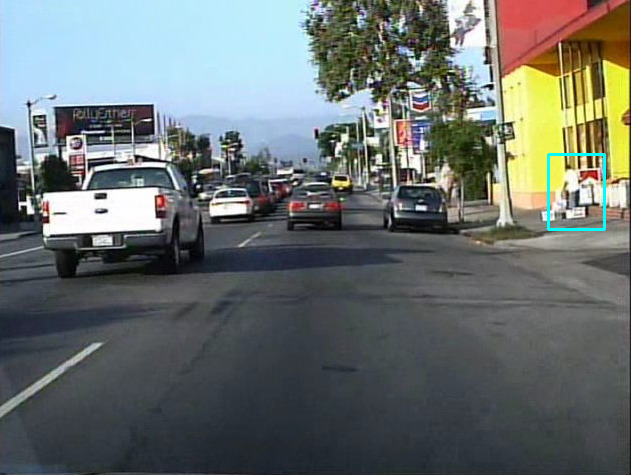} \hfill \includegraphics[width=1.5in]{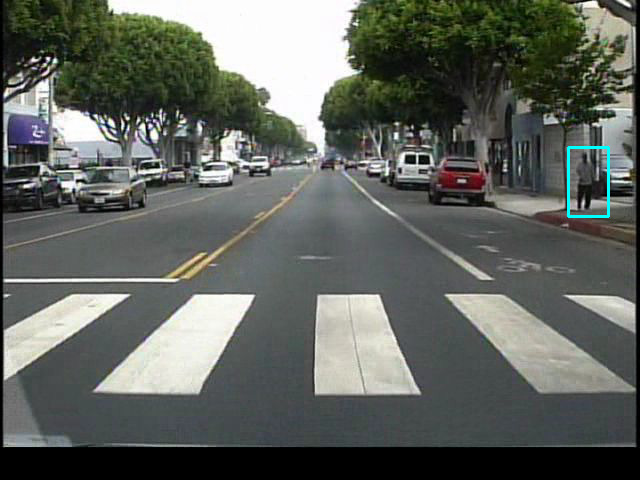}  \hfill\\
          Variation in lighting condition \hfill \\
          \includegraphics[width=1.5in]{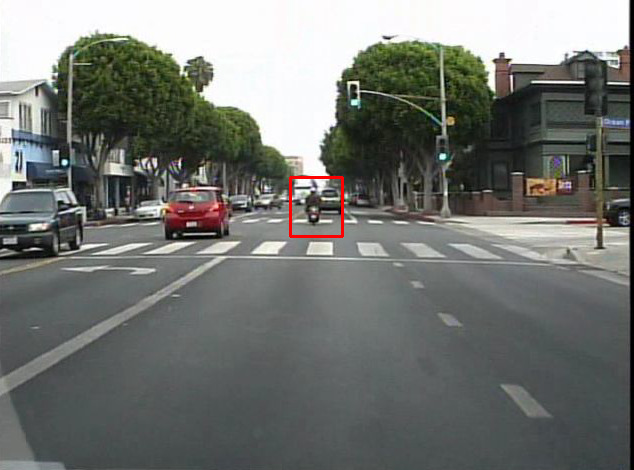} \hfill \includegraphics[width=1.5in]{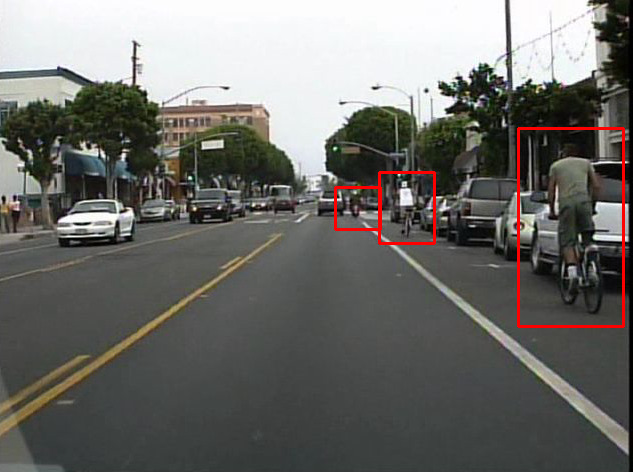}  \hfill\\
          Variation in pedestrian size  \hfill \\
          \includegraphics[width=1.5in]{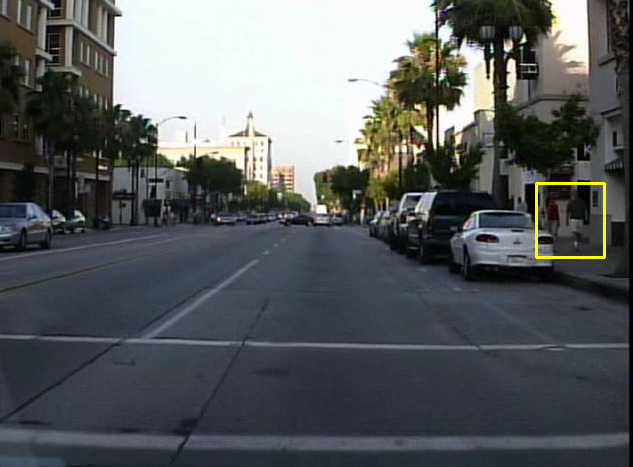} \hfill \includegraphics[width=1.5in]{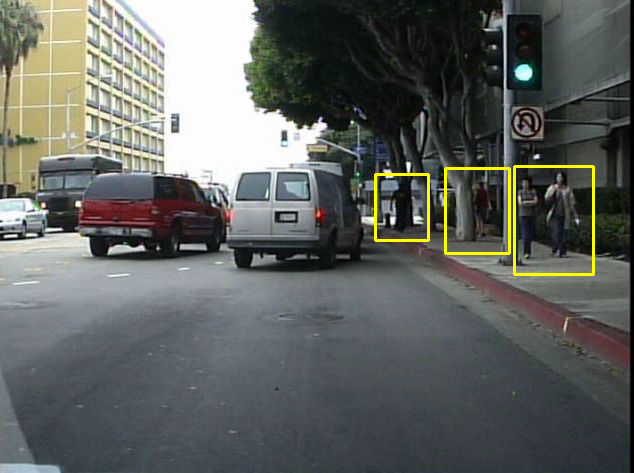}  \hfill\\
          Variation in occlusion by human body \hfill \\
          \includegraphics[width=1.5in]{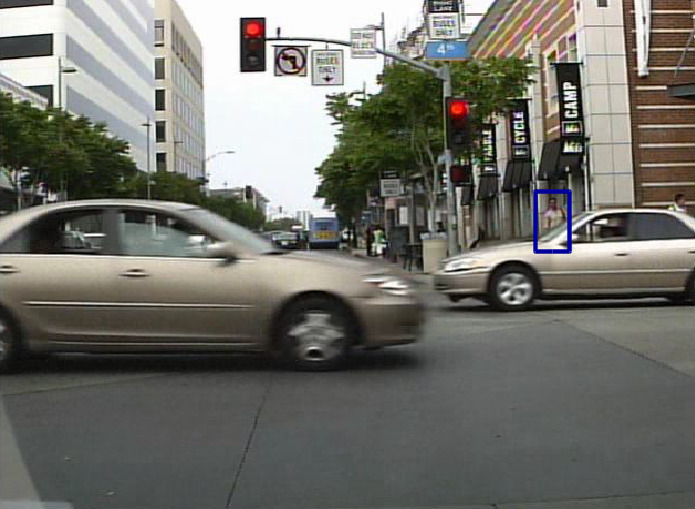} \hfill \includegraphics[width=1.5in]{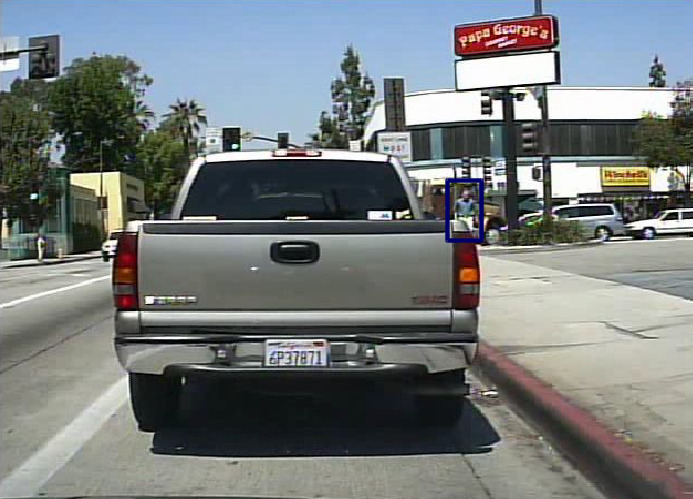}\hfill\\
           Variation of occlusion by non-human body \hfill \\
  \caption{A few examples of the challenges of existing Caltech Pedestrian Dataset}       \label{fig:challenges}
 \end{center}
   \end{figure}
    Also detecting pedestrians in a dynamic background (camera set in a moving car) is more difficult than detecting the same in a static background (fixed traffic camera). It is noteworthy that the resolution of the camera is also an important factor to accurately detect a pedestrian. Many traditional and deep learning based approaches are exploited to find a generic solution of this problem.
    \item \textbf{Real time detection} of pedestrian is an even more challenging task. But this is required if we need to incorporate the detection algorithm in a real life system. This is usually measured by Frames per Second (FPS). It is very challenging to implement a network that produce results with a very low MR (high recognition accuracy, for an ideal system this should be 0) and at the same time very high FPS. It is obvious that higher recognition accuracy needs more processing \textit{time} per frame which on the other hand decreases FPS. This trade-off can be managed to some extent by providing large computational resources which is not available in general purpose systems.
\end{itemize}
In the present work we have taken care of both these issues and compared the performance of our model with respect to FPS and MR to other benchmark results.

\section{Related Works}
\label{sec:relatedwork}

The relevant studies in this field aim for optimizing  miss-rate and real time processing performance. A number of strategies of pedestrian detection have been proposed in the literature. But it is hard to improve both miss-rate and processing speed simultaneously. SVM (Support Vector Machine) trained on HOG (histogram oriented gradients) features is a popularly chosen classifier and some significant studies under this framework had been reported in \cite{dollar2012pedestrian}, \cite{alonso2007combination} and \cite{xu2005pedestrian}. Some other similar studies were reported in \cite{tuzel2008pedestrian} and \cite{dalal2005histograms} as well. Benenson et al. \cite{benenson2012pedestrian} proposed a method without using any deep learning architecture for real time processing which could process 135 frames per second but the miss-rate was noted to be as high as 42\%. Ding et al. and Xiao et al. had used certain Contextual Boost method \cite{ding2012contextual} with the help of an AdaBoost Classifier \cite{freund2001adaptive} to improve the performance of pedestrian detection based on contextual information and achieved a miss-rate of 25\%. Convolutional neural network (CNN) based classifiers have already established their efficiency in pedestrian detection tasks \cite{luo2014switchable} \cite{ren2015faster} \cite{angelova2015real}. Vanilla Faster R-CNN had performance limitation on Caltech pedestrian dataset due to smaller sizes of pedestrians. Later, He et al. improved this Faster R-CNN strategy with the help of the (RPN + BF\cite{appel2013quickly}) \cite{zhang2016faster} to reduce the miss rate but it could process only 2 frames per second with a miss-rate of 9.6\% which is the present state-of-the-art of pedestrian detection. Several deep learning based models have been proposed to speedup the processing. A recent approach introduced by Angelova et al. \cite{angelova2015real} could process at 15 FPS but the miss-rate was reported to be 26.21\%.  \\

During the last decade a number of video / image datasets of pedestrian samples have been made available for research purposes. MIT pedestrian dataset consisting of 924 instances of pedestrians was first introduced by Papageorgiou et al. \cite{papageorgiou2000trainable} towards initiation of systematic studies of pedestrian detection. Some of the well-known pedestrian datasets include INRIA \cite{dalal2005histograms}, Daimler \cite{enzweiler2008monocular}, ETH \cite{ess2008mobile}, PPSS \cite{luo2013pedestrian} etc. The two widely popular datasets containing large number of samples include Caltech Pedestrian dataset \cite{dollar2009pedestrian} and KITTI dataset \cite{geiger2012we}. Recently, Zhang et al. \cite{zhang2017citypersons} has introduced Citypersons, a pedestrian dataset consisting of a diverse set of stereo video sequences recorded in
streets of different cities.

\section{Development of a New Pedestrian Dataset} \label{sec:datasetpreparation}

\begin{figure}[]
    \includegraphics[width=2.5in, height=2.5in]{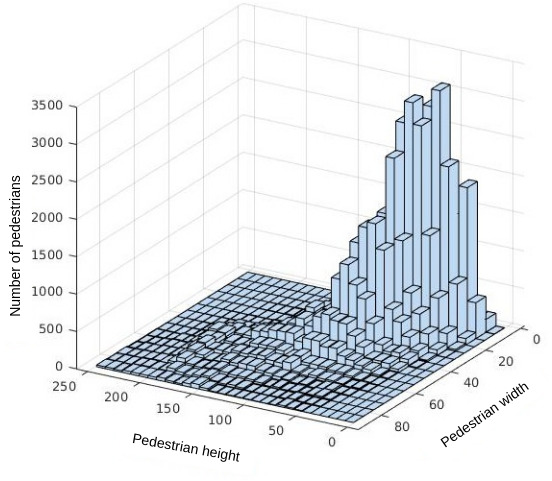}
  \caption{Histogram of height and width of bounding boxes of pedestrians of the present sample database} \label{fig:dataset_distribution}
\end{figure}

As mentioned in Sec.\ref{sec:intro}, usually a large volume of training samples is required for proper training of a CNN based deep network. Also, the training samples should contain sufficient variations with respect to different factors as described before. Therefore, we took an initiative to develop a new sample dataset for supplementing the existing popular Caltech Pedestrian Dataset (CPD). We call this new dataset ISI Pedestrian Dataset (ISIPD) and the same will be freely distributed on request from academic researchers. This new ISIPD contains 13,129 annotated video frames and its annotation includes 82.3K pedestrian bounding boxes. The distribution of height and width of these bounding boxes are shown in Fig. \ref{fig:dataset_distribution}.

\begin{figure}[h]

\includegraphics[width=1.5in]{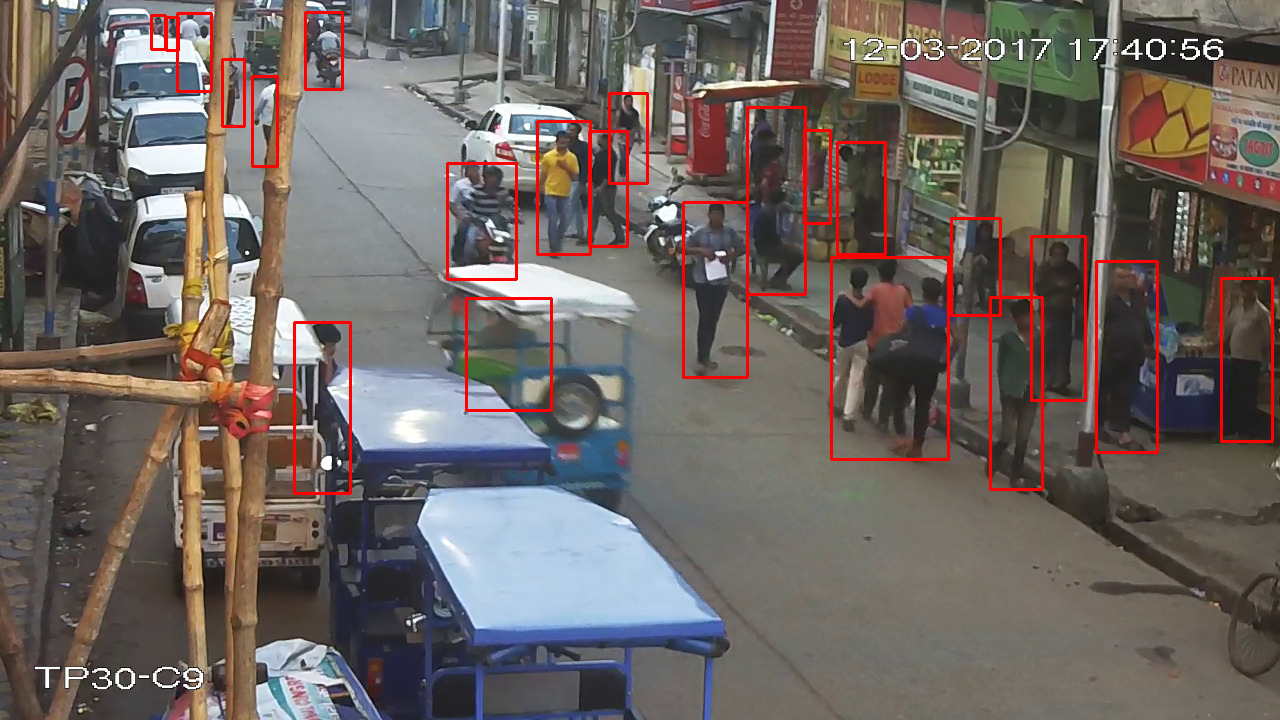}\hfill
\includegraphics[width=1.5in]{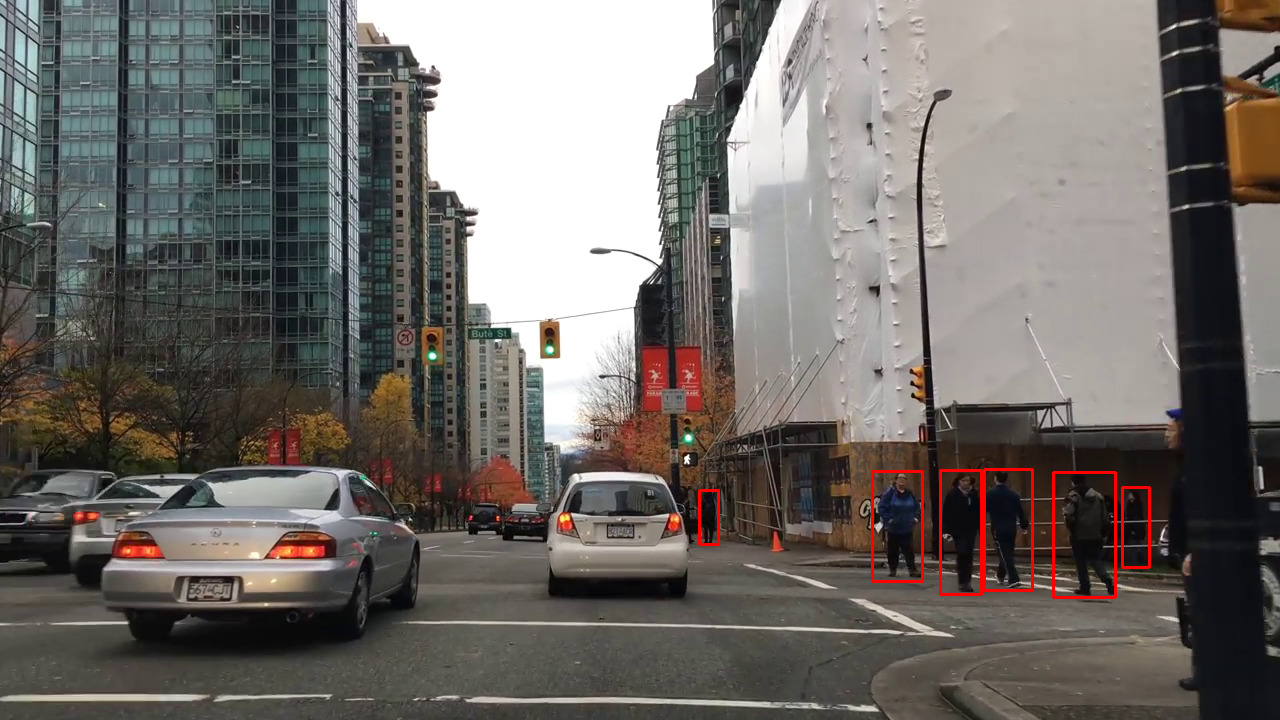}\\
\includegraphics[width=1.5in]{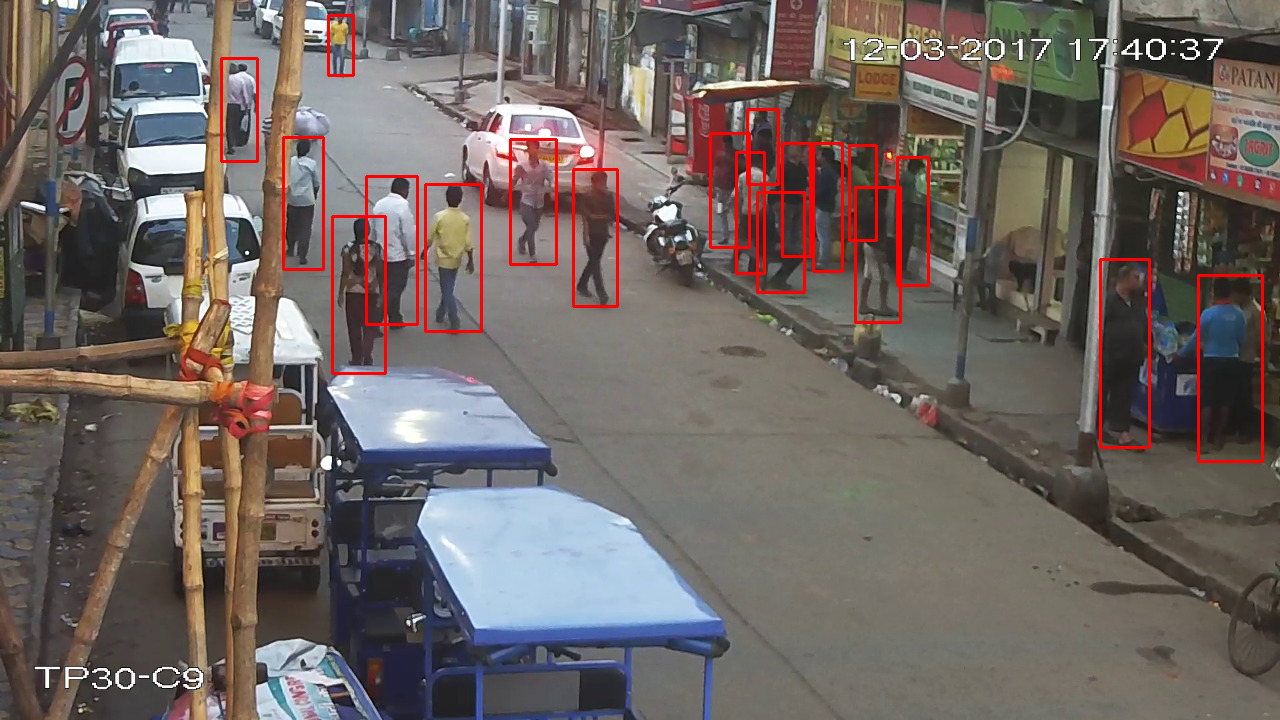}\hfill
\includegraphics[width=1.5in]{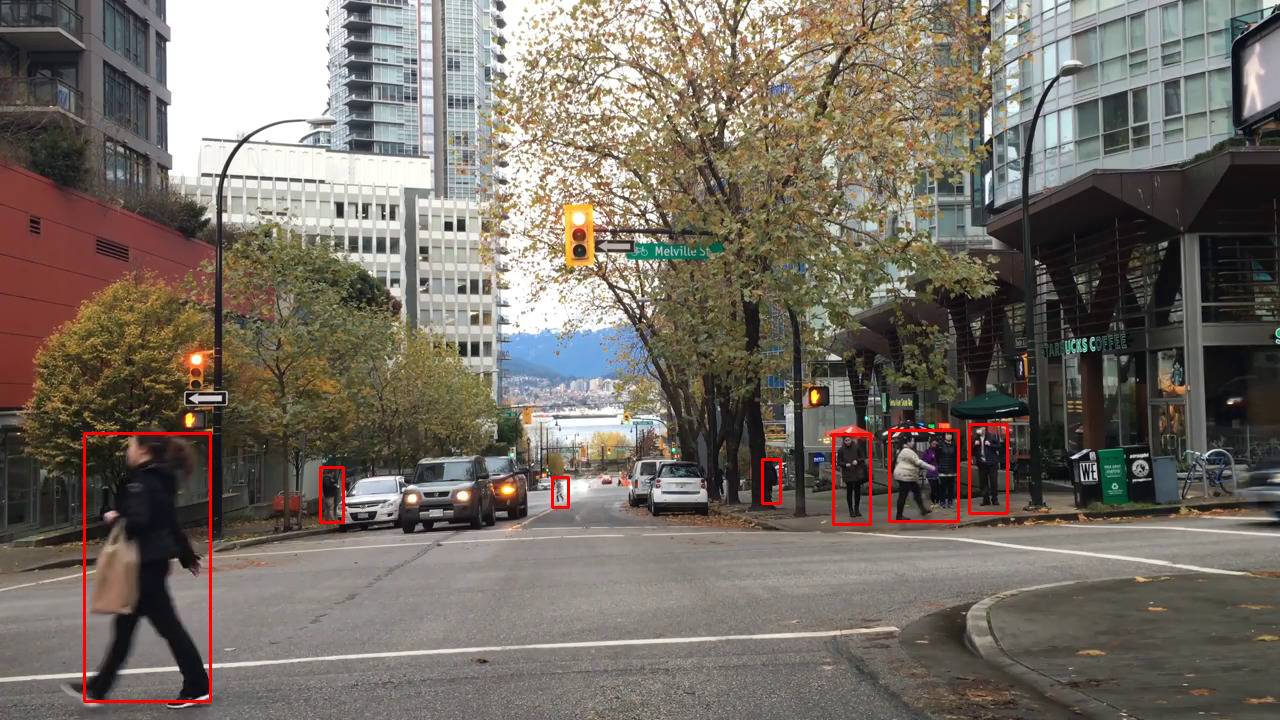}\\
\includegraphics[width=1.5in]{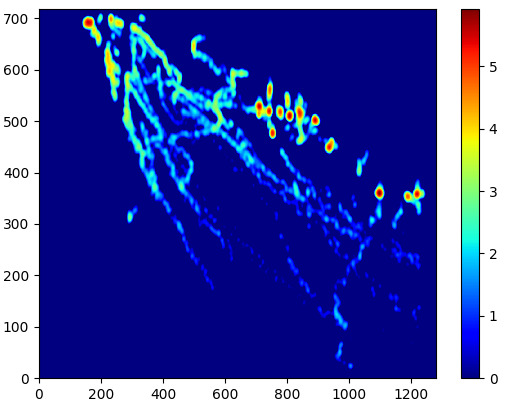} \hfill
\includegraphics[width=1.5in]{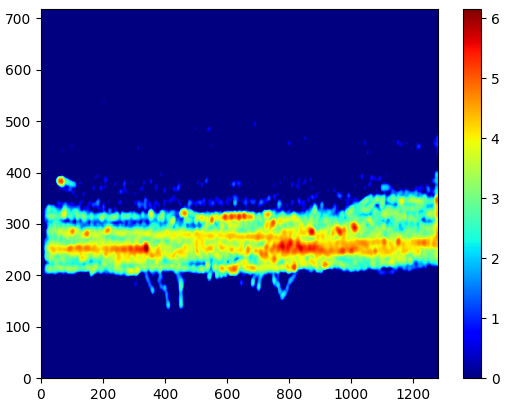}\\
\caption{Two columns shows two different camera angles, traffic conditions and pedestrian variations. Corresponding distribution heat map is given in last row of each column.}
\label{fig:distribution}
\end{figure}

The resolution of the video frames of CPD is $640 \times 480$ pixels whereas the same of ISIPD is $1280 \times 720$ pixels. We have kept the resolution higher to facilitate the detection of smaller pedestrian figures. Several video segments captured by different cameras fitted in different vehicles and a number of traffic surveillance cameras under normal traffic conditions of different urban areas of India, North America and Germany have been used to develop this new dataset. The video segments are so selected that various possible characteristics of the crowd or pedestrians get represented in this new dataset. On the contrary, the existing CPD consists of approximately 10 hours of continuous video taken from a particular vehicle driving through regular traffic of certain urban environment. There are video frames in the CPD which do not have any pedestrian. On the other hand, each video frame of ISIPD has at least one pedestrian. The pedestrian bounding boxes of each image of ISIPD has been manually annotated using a tailor made software tool. In Fig. \ref{fig:distribution}, two different traffic scenario is presented along with the heat-map of log-normalized distribution of the center positions of pedestrians.

\begin{figure}[H]
        \includegraphics[height=.9in, width=.4in]{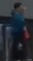}
        \includegraphics[height=.9in, width=.4in]{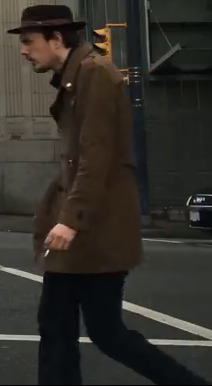}
        \includegraphics[height=.9in, width=.4in]{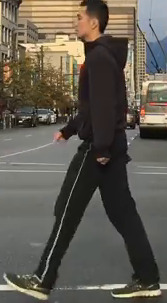}
        \includegraphics[height=.9in, width=.4in]{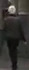}
        \includegraphics[height=.9in, width=.4in]{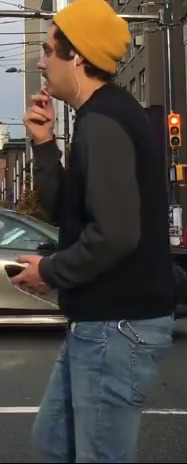}
        \includegraphics[height=.9in, width=.4in]{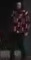}
        \includegraphics[height=.9in, width=.4in]{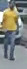} \\
        \includegraphics[height=.9in, width=.4in]{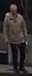}
        \includegraphics[height=.9in, width=.4in]{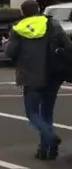}
        \includegraphics[height=.9in, width=.4in]{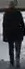}
        \includegraphics[height=.9in, width=.4in]{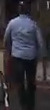}
        \includegraphics[height=.9in, width=.4in]{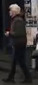}
        \includegraphics[height=.9in, width=.4in]{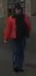}
        \includegraphics[height=.9in, width=.4in]{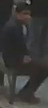} \\
      \caption{A few pedestrian samples of ISIPD representing its wide variations with respect to clothing, lighting condition, pose, resolution etc.} \label{fig:dataset_example}
\end{figure}

The training set of existing CPD consists of 128K images and 192K pedestrian bounding boxes. The development of new ISIPD is aimed at capturing more variations in the training samples towards efficient training of the network. A few pedestrian image samples of ISIPD are shown in Fig. \ref{fig:dataset_example}.

Volumes of training sets of different sample databases used in the present study are shown in Table \ref{tab:statistics}.

\begin{table}[H]
\caption{Training set size of different sample datasets used in the present study} \label{tab:statistics}
\scalebox{.8}{%
\begin{tabular}{|c|c|c|c|c|}
\hline
Dataset Name & Total Frames & BBoxs &   Resolution\\
\cline{3-4}
\hline
CPD &  128K & 192K &  $ 640 \times 480 $\\
\hline
CityPersons \cite{zhang2017citypersons} & 2.9K & 19.6K &  $ 2048 \times 1024 $\\
\hline
ISIPD & 13.1K & 82.3K & $ 1280 \times 720 $\\
\hline
\end{tabular}}
\end{table}

\section{Proposed Model}
\label{sec:proposedmodel}

An important feature of the proposed model is its simplicity. In the proposed strategy, we first look for \textit{potential zones} (Seek) where pedestrian figures may exist. Such a potential zone is an area of the frame where the following situations may occur,
\begin{itemize}
    \item It may contain pedestrian figures of very small size (due to its distance from the camera).
    \item It may contain pedestrian figures of very large size (due to its proximity to the camera). A few such figures may be so large that it may be distributed over multiple adjacent regions.
    \item Multiple pedestrians may appear in a single region. Some of these pedestrian figures may appear connected.
\end{itemize}

\begin{figure}[!h]

	\includegraphics[width=1.6in]{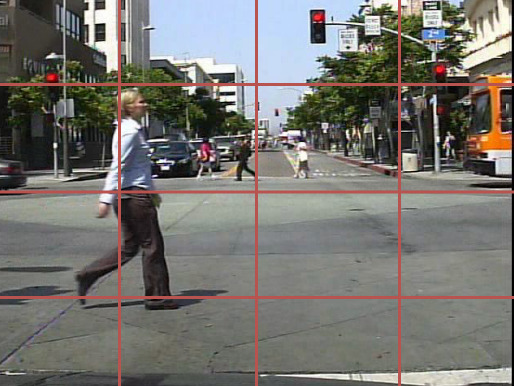}
	\includegraphics[width=1.6in]{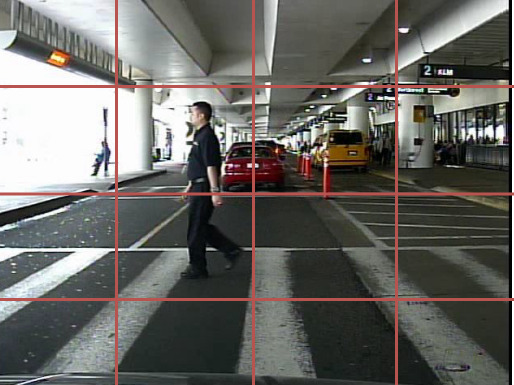}\\
	\vspace{-0.18in}
	\caption{4$\times$4 grid views of two original image frames. The frame to the left has eight positive sub-images while the frame to the right has three positive sub-images. }
	\label{fig:tiny_classifier_data_grid}
\end{figure}

\begin{figure}[]
	\includegraphics[width=.8in, height=1in]{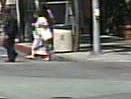}
	\includegraphics[width=.8in, height=1in]{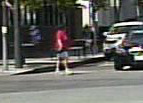}
	\includegraphics[width=.8in, height=1in]{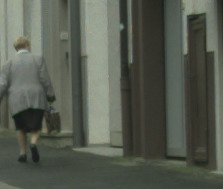}
	\includegraphics[width=.8in, height=1in]{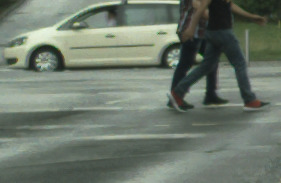}
	\includegraphics[width=.8in, height=1in]{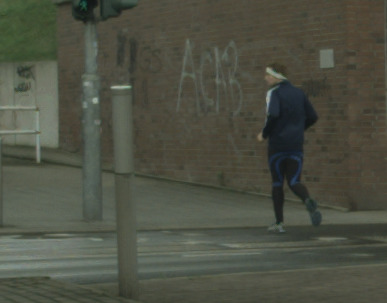}
	\includegraphics[width=.8in, height=1in]{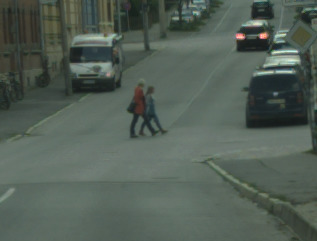}
	\includegraphics[width=.8in, height=1in]{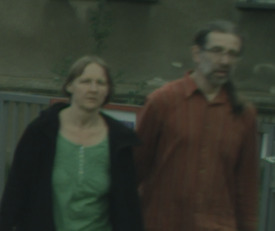}
	\includegraphics[width=.8in, height=1in]{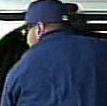}\\
	\vspace{-0.15in}
	\caption{A few positive training samples for classifier $C_z$.}
	\label{fig:data_tiny_network_positive}
\end{figure}

Once a zone is identified as a potential one, we further scan it (using a sliding window) to compute the bounding box of each pedestrian or connected pedestrian(s) (Find). We name this architecture as \textbf{Seek and You will Find (SaYwF)}. The advantage of this strategy is that the parts of the frame which are unlikely to have a pedestrian figure can be rejected as non-potential zone. Thus we can restrict the number of expensive sliding window operations and subsequently improve the overall FPS. The only issue that remains is the correct identification of a potential zone. Errors in this detection stage may drop the MR heavily. We have solved the problem by introducing a simple feed-forward multi-layer classifier that can distinguish between potential and non-potential zones. We term this classifier as \textbf{Zone Classifier ($C_z$)}. It is based on a CNN based Inception style network \cite{Szegedy_2015_CVPR} details of which have been discussed in Sec. \ref{sec:custom_inception}. Training of $C_z$ is accomplished by dividing each training frame image in $4\times4$ grids (as shown in Fig. \ref{fig:tiny_classifier_data_grid}) yielding $16$ sub-images from each frame. The annotation groundtruth is consulted to label each such sub-image. If \textit{any} part of a pedestrian figure falls within a sub-image, it is considered as a \textbf{positive} sample and otherwise the same is treated as a \textbf{negative} sample. Some positive training samples are shown in Fig. \ref{fig:data_tiny_network_positive}. It may be noted that some of the positive samples contain multiple pedestrians or only a part of a pedestrian.

Once the zone classifier is ready the detection can be done in three phases.  In the first phase we use ($C_z$) to identify an area of the frame as a potential zone. As we have no prior information about the number, size and position of pedestrian figures, we again sub divide original frame by a grid (yielding multiple sub-images). Feeding each one of the sub-image to ($C_z$) result confidence score of that region. Based on these scores we select a region for further processing. This is explained in Sec. \ref{sec:phaseII}.\par
In the second phase, we need to find the exact bounding box around each pedestrian from the potential zones suggested in phase I. To achieve this we trained another deeper binary classifier (referred as \textbf{Pedestrian Classifier or $C_p$}) based on the similar architecture as said earlier, that identifies the pedestrian and this trained classifier is fed with regions taken from a sliding window over a potential zone. Note that the sliding window size and stride are two important parameters to correctly find a pedestrian figure.\par
In the third phase, we used a Non-Maximum Suppression (NMS) method to reject multiple bounding box suggestions for same pedestrian and finally draw the bounding box. Instead of densely scanning all possible regions of an input video frame, we densely scan only proposed potential zones.

\subsection{Modified Inception Architecture }
\label{sec:custom_inception}

\begin{figure}[h]
        \includegraphics[width=2.5in, height=2.5in]{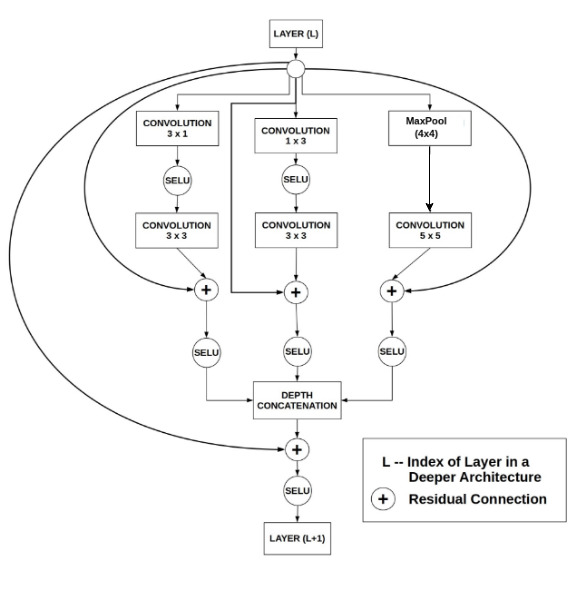}
        \vspace{-0.2in}
        \caption{Inception style CNN architecture}
        \label{fig:inception_layer}
\end{figure}

We propose a densely connected CNN based modified Inception architecture to build a classifier. One such inception block is shown in Fig. \ref{fig:inception_layer}. In our architecture we have appended multiple such blocks sequentially. This is illustrated in Fig. \ref{fig:tiny_classifier} and Fig. \ref{fig:full_classifier}. A few important attributes of the inception block includes the following:

\begin{itemize}
    \item Two filters having different orientation is used. One of them is horizontal ($ 1 \times 3 $) and the other one is vertical ($ 3 \times 1 $). Instead of using a traditional ($3 \times 3$) filter, we have used these two asymmetric filters to reduce the computation time. Number of multiplications in our scheme with two filters are $6$. Whereas the same for a ($3\times3$) filter is 9. As the computation time is proportional to number of multiplications we achieve a 1.5 times faster computation in our scheme.
    \item First layer of our inception block is to reduce the dimension (channels only) of the input coming from previous layers. This reduced feature map is processed by an expensive (3 $ \times $3) and (5 $\times $ 5) filters followed by a max pooling layer with (4 $ \times $4) with 1 stride.
    \item A residual connection to reuse the original context of the information from previous layer is used. This also prevents the problem of vanishing or exploding gradient during gradient descent optimization as mentioned in \cite{he2016deep}. It also reduces the time to convergence than the traditional Inception network \cite{huang2017densely}.
    \item Smaller filters ($ 1 \times 3 $) are used prior to larger filters ($3 \times 3$) as smaller filters preserves the special context and larger filters extract higher dimensional features.
\end{itemize}

We incorporated Batch Normalization (BN) layer as a regularizer \cite{ioffe2015batch} which helps for faster learning. In the original paper \cite{he2016deep}, He et al.  applied BN in $Convolution \rightarrow BN \rightarrow ReLU $ order.
\par Consider x is feature vector, W is weight matrix and b is the bias,

\begin{eqnarray}
z  \leftarrow ReLU(BN(Wx + b)) \nonumber \\   y  \leftarrow BN(Wx + b)
\end{eqnarray}
\begin{eqnarray}
    E[z] = E[ReLU(y)] > E[y] = 0, \nonumber \\ Var[z] = var[Relu(y)] < var[y] = 1
\end{eqnarray}

 But the variance and mean of normalized output from $BN$ layer is altered by the ReLU \cite{krizhevsky2012imagenet} operation as ReLU function transforms all the negative value to positive. For this reason, In our architecture we have used Scaled Exponential Linear Unit (SELU) \cite{klambauer2017self} activation function (refer to Eqn. \ref{eqn:selu}) after Convolution to avoid this inconsistency.

\begin{equation}
    SELU(x) = \lambda \begin{cases}
        x & \text{if $x>0$}\\
        \alpha^x - \alpha & \text{if $x \leq 0$}
    \end{cases}
    \label{eqn:selu}
\end{equation}

where $\lambda$ and $\alpha$ are fixed parameters. The negative and positive value output of the Convolution layer remain same in this SELU activation function to control the zero mean and unit variance. It helps for faster learning because of approximately zero mean and avoid the vanishing/exploding gradient problem.

\subsection{Phase I: Seek if any pedestrian is there}
\label{sec:phaseI}
\begin{figure}[!h]
	\includegraphics[width=3in, height=1.2in]{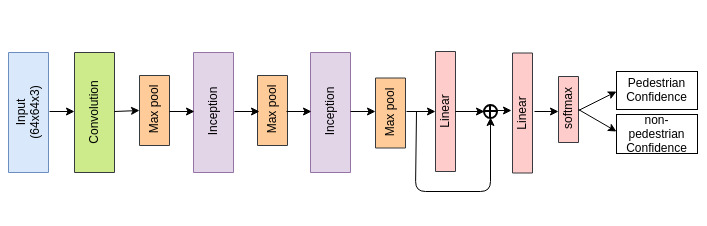}
	\vspace{-0.2in}
	\caption{Potential Zone Classifier $C_z$}
	\label{fig:tiny_classifier}
\end{figure}

As mentioned earlier this phase consists of a binary classifier ($C_z$) that can classify a region of the image as potential zone. We have used our modified Inception block to build this classifier. This is shown in Fig. \ref{fig:tiny_classifier}. Size of the input RGB color image is $64\times64$. Details of the classifier network is described bellow.

The frame is divided into $16$ regions by a $4 \times 4$ grid. Each one of these regions yields a confidence score as a potential zone when given as input to our zone classifier $C_z$. If a region is not a potential zone (there are no pedestrian or part of pedestrian in this region) it is discarded, otherwise it is marked by a special flag and passed to Phase II.
 We have sampled 100K positive images and 180k negative images to train $C_z$.

\begin{figure}[]
	\includegraphics[width=.5in, height=.9in]{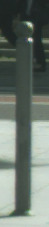}
	\includegraphics[width=.5in, height=.9in]{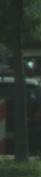}
	\includegraphics[width=.5in, height=.9in]{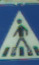}
	\includegraphics[width=.5in, height=.9in]{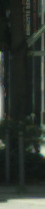}
	\includegraphics[width=.5in, height=.9in]{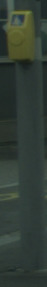}
	\includegraphics[width=.5in, height=.9in]{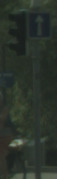}\\
	
	\includegraphics[width=.5in, height=.9in]{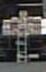}
	\includegraphics[width=.5in, height=.9in]{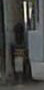}
	\includegraphics[width=.5in, height=.9in]{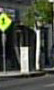}
	\includegraphics[width=.5in, height=.9in]{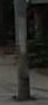}
	\includegraphics[width=.5in, height=.9in]{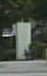}
	\includegraphics[width=.5in, height=.9in]{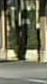}
    \vspace{-0.15in}
	\caption{A few hard-negative training samples.}
	\label{fig:hard_negative_data}
\end{figure}

\begin{table*}[!h]
\centering
\caption{Details of the Zone Classifier.} \label{tab:classifier_network}
\scalebox{0.8}{%
\begin{tabular}{|c|c|c|c|c|c|c|c|c|c|c|c|c|}
\hline
Type & Patch size/stride & Output Size & \# $1\times1$ residual & \#$1\times3$  & \#$3\times3$ & \#$1\times1$ residual & \#$3\times1$ & \#$3\times3$ & \#residual $1\times1$ & $5\times5$ & bridge residual  & params\\
\hline\hline
convolution & 3$\times$3/2& $32\times32$/32  &- &- & -& - &- &- &- & - &- &896\\[2ex]
\hline
max Pool & 4$\times$4/2 & 16$\times$16/32 & - &- &- &- &- &- &- &- & - &-\\[2ex]
\hline
inception &- & 16$\times$16/64  & 16 & 8 & 16 & 16& 8 & 16 & 32   & 32& 64 & 33744\\[2ex]
\hline
max Pool & 4$\times$4/2 & 8$\times$8/64 & -& -&- &- &- & -& -&- & -&- \\[2ex]
\hline
inception & - & 8$\times$8/96 & 32 & 16 & 32 & 32& 16 & 32 & 64   & 64& 128 & 79168\\[2ex]
\hline
max Pool & 4$\times$4/2 & 4$\times$4/96  &- &- &- &- &- &- &- &- &-   &-\\[2ex]
\hline
linear &- & 128 & -  & -&  -&  -&  -& - &- &- &-  &196736\\[2ex]
\hline
residual & -& 128 & -   &- & - &-  & - & - & -&-  &- &196736\\[2ex]
\hline
linear & -& 2 & - & -  & - & - & - &-  &- &- & - &514\\[2ex]
\hline
softmax & - & 2 &-  &- & - & - & - & - &- &-  & - &-\\[2ex]
\hline
\end{tabular}}
\end{table*}

\begin{table*}[!h]
\centering
\caption{Details of the Pedestrian Classifier.} \label{tab:classifier_network}
\begin{center}
\scalebox{0.8}{%
\begin{tabular}{|c|c|c|c|c|c|c|c|c|c|c|c|c|}
\hline
Type & Patch size/stride & Output Size & \# $1\times1$ residual & \#$1\times3$  & $\#3\times3$ & \#$1\times1$ residual & \#$3\times1$ & \#$3\times3$ & \#residual $1\times1$  & $5\times5$ & bridge residual  & params\\
\hline\hline
convolution & 3$\times$3/1& 64$\times$64/32  &- &- & -& -&- &- &- &  -  &- &896\\[2ex]
\hline
max Pool & 4$\times$4/2 & 32$\times$32/32 & - &- &- &- &- &- &- &- & - &-\\[2ex]
\hline
inception &- & 32$\times$32/64  & 32 & 16 & 32 & 32& 16 & 32 & 16  & 16 & 64& 24080\\[2ex]
\hline
max Pool & 4$\times$4/2 & 16$\times$16/64 & -& -&- &- &- & -& -&-  & -&- \\[2ex]
\hline
inception &- & 16$\times$16/128 & 48 & 32 &48 & 48 & 32 & 48 & 32 &  32& 128 & 91584\\[2ex]
\hline
inception &- & 16$\times$16/128 & 48 & 32 &48 & 48 & 32 & 48 & 32 & 32& 128 & 111040\\[2ex]
\hline
max Pool & 4$\times$4/2 & 8$\times$8/128  &- &- &- &- &- &- &- &- &-   &-\\[2ex]
\hline
inception &- & 8$\times$8/200 &  80 & 40 &80 & 80 & 40 & 80 & 40 &  40& 200 & 226320\\[2ex]

\hline
inception &- & 8$\times$8/200 &  80 & 40 &80 & 80 & 40 & 80 & 40 &  40& 200 & 226320\\[2ex]

\hline
max Pool & 4x4/2 & 4x4/200 & - &- &- &- &- &- &-  & -& - &-\\[2ex]
\hline
linear &- & 512 & -  & -&  -&  -&  -& - &- &- &- &1638912\\[2ex]
\hline
linear & -& 256 & -  &- &-  &-  &-  &-  & -& - & - &131328\\[2ex]
\hline
residual & -& 256 & -   &- & - &-  & - & - & - &-  &- &819456\\[2ex]
\hline
linear & -& 2 & - & -  & - & - & - &-  &-  & -& - &514\\[2ex]
\hline
softmax & - & 2 &-  &- & - & - & - & - &- &- & - &-\\[2ex]
\hline
\end{tabular}}
\end{center}
\end{table*}

\subsection{Phase II: Find where are the pedestrians} \label{sec:phaseII}
\begin{figure*}[!h]
	\includegraphics[width=\textwidth]{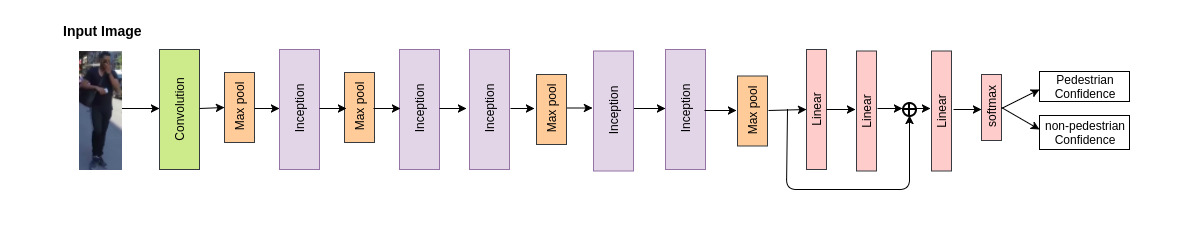}
	\caption{Pedestrian Detection Classifier $C_p$}
	\label{fig:full_classifier}
\end{figure*}

The classifier $C_z$ of Phase I has already provided  a collection of potential zones $Z_1, Z_2, \ldots$. Now, a sliding window of size 16 $\times$ 16 is densely moved (small step size) over each $Z_i$. Use of a larger step size should plummet the MR while it may cause increased FPS. Here, a step size of 5 has been empirically selected. The effect of step size on MR and FPS has been discussed in Sec. \ref{sec:experimentalresult}. The region or area cropped by the sliding window is re-scaled to feed the same as input to the pedestrian detection classifier $C_p$ (Fig. \ref{fig:full_classifier}).

\begin{figure}[h]
    \includegraphics[width=2.5in,height=3in]{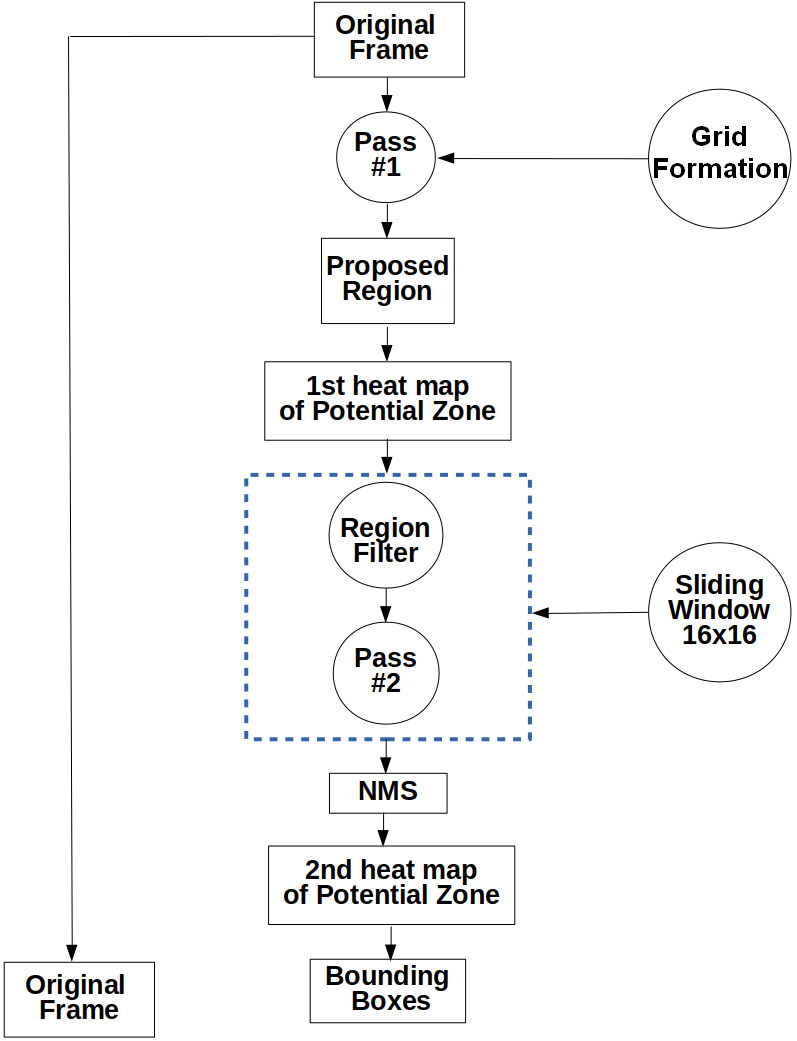}
    \vspace{-.15in}
  \caption{Flow chart of the proposed method.}
  \label{fig:sliding_window}
  \vspace{-0.20in}
\end{figure}

A detail configuration of the classification model is given in Table \ref{tab:classifier_network}.
Total number of parameters of our deep architecture is $\sim$3.3 Million which is much less than Deep Network Cascades (DNC) which consists of $\sim$50 million parameters. The processing as done in Phase I and II are shown graphically in Fig. \ref{fig:sliding_window}.

\subsection{Phase III: One and only one}
\label{sec:seek_find}

The image region covered in one sliding window position within a potential zone is processed by the classifier ($C_p$) trained previously (refer to Sec. \ref{sec:phaseII}). When the scan is complete we get a heatmap of the frame denoting existence of pedestrians (detection window). It should be noted that after the sliding window completes the scan of a single potential zone. We may get multiple overlapping detection windows inside a potential zone for a single pedestrian position. Thus redundant bounding boxes are eliminated by Non-Maximum Suppression (NMS). It is done by not considering image regions (corresponds to different sliding window positions) that are overlapping with some detection window by 50\% IOU (Intersection over Union).


\par
\begin{figure}[!h]

    \centering
    
    \includegraphics[width=1.5in]{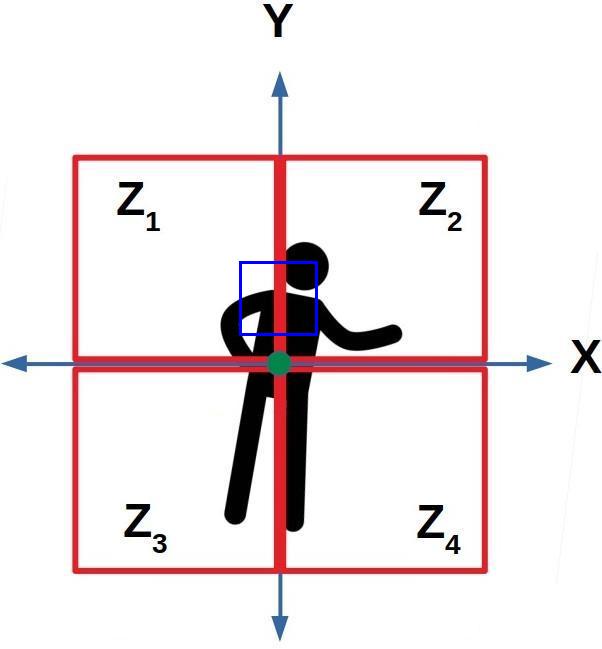}
    \caption{A single pedestrian figure is distributed over four adjacent regions each of which has been identified as a potential zone by the classifier  $C_z$. The sliding window (blue) centered on the boundary of $Z_1$ and $Z_2$ has significant overlap with pedestrian bounding boxes of both of the two zones.}
    \label{fig:multiple_positive}
    \vspace{-0.12in}
\end{figure}

 Now we face an interesting situation when a pedestrian figure is distributed over multiple adjacent potential zones. One such scenario is shown in Fig. \ref{fig:multiple_positive}, where $C_z$ has detected $Z_1$, $Z_2$, $Z_3$ and $Z_4$ as potential zones. $C_p$ has estimated the bounding boxes (say, $W_1$, $W_2$, $W_3$ and $W_4$) for individual parts of the pedestrian lying in each of these zones. Our task is to identify the occurrence of a similar situation and merge the corresponding bounding boxes. Towards the same, we first detect pairs of adjacent potential zones. For each such pair, we run a sliding window ($S_w$) of size $32\times32$ with its center on the common boundary. $S_w$ is resized to $64\times64$ to feed it to the classifier $C_p$. If $S_w$ is classified as pedestrian, then we further compute its overlaps with pedestrian bounding boxes (if any) of both the adjacent zones. If at least one such $S_w$ has significant overlap with a pedestrian bounding box lying in each of the two zones, then the corresponding pedestrian bounding boxes of the two zones are merged.

\section{Experimental Results} \label{sec:experimentalresult}
There are many interesting aspects related to pedestrian detection problem using CNN based deep learning methods. Some of them include (i) the trade off between FPS and MR, (ii) effect of stride size on FPS and MR, (iii) effect of selection of training and test sets etc. We have explored quite a few of these aspects through experiments.\par
As mentioned earlier, our model uses two classifiers $C_z$ (Fig. \ref{fig:tiny_classifier}) and $C_p$ (Fig. \ref{fig:full_classifier}). These two were trained using training sets $D_z$ and $D_p$ respectively which were formed from training samples of three datasets viz., CPD, CityPersons and ISIPD. Training and test set details including the numbers of positive and negative samples have been provided in Table \ref{tab:classifier_dataset}. Training was executed on a system consisting of three Nvidia Tesla P6 GPU and for each mini batch of size 2048, it takes ~700 milliseconds. Training is continued until training loss is saturated.
\begin{table}
\caption{Dataset ($D_z$ and $D_p$) Details.}
\vspace{-0.12in}
\begin{tabular}{|c |c | c| c| c |}
 \hline
 Model & \multicolumn{2}{c|}{Training set}  & \multicolumn{2}{c|}{Test set}  \\ [0.5ex]
 \hline
  Type of Sample & Positive & Negative & Positive & Negative \\
  \hline
 $D_z$ & 100k & 180k & 50k & 50k \\
 \hline
 $D_p$ & 210k & 1.2 million & 90k & 90k \\
 \hline
\end{tabular}
\label{tab:classifier_dataset}
\vspace{-0.2in}
\end{table}

\par The performance of the proposed end-to-end pedestrian detection system has been measured based on MR and FPS. MR is computed by comparing the IOU of bounding boxes from original annotated image with the predicted bounding boxes. If the overlap between actual and predicted bounding box is more than 50\% then we consider this as a hit. Miss can occur if either we positively detect a region with out pedestrian or fail to detect a region with pedestrian. Miss rate is computed based on these outcome. Our model can process 20 frames on Nvidia Titan XP GPU. To compare the performance of our model is compared with some existing state of the art deep CNN based architectures. Note that only those models that can process at least 10 frames per second (10 FPS and above) are chosen.

\begin{figure}[!h]
\vspace{-0.12in}
\begin{tabular}{cc}
   \includegraphics[width=2in]{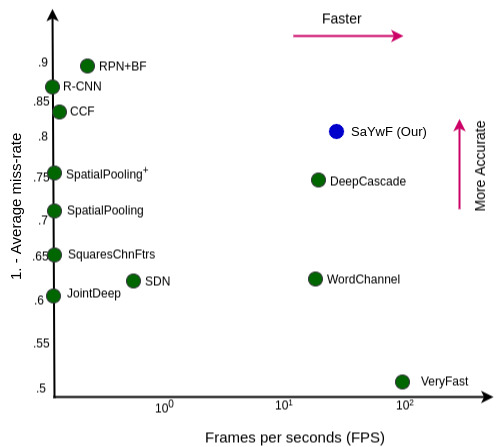}
\end{tabular}

\vspace{-0.18in}
\caption{Performance of different model on Caltech Pedestrian Dataset.}

\label{fig:stride_plot}
\vspace{-0.25in}
\end{figure}

\begin{figure}[!h]
    \includegraphics[width=1.5in,height=1in]{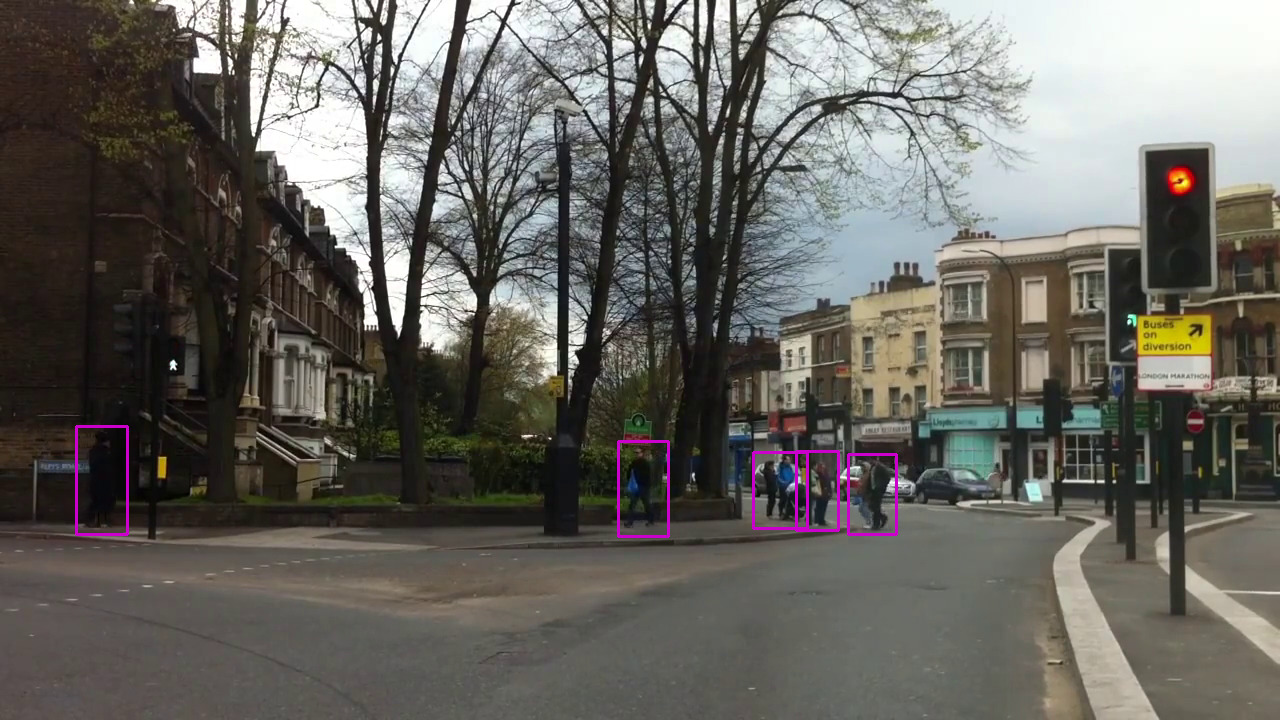} \hfill \includegraphics[width=1.5in,height=1in]{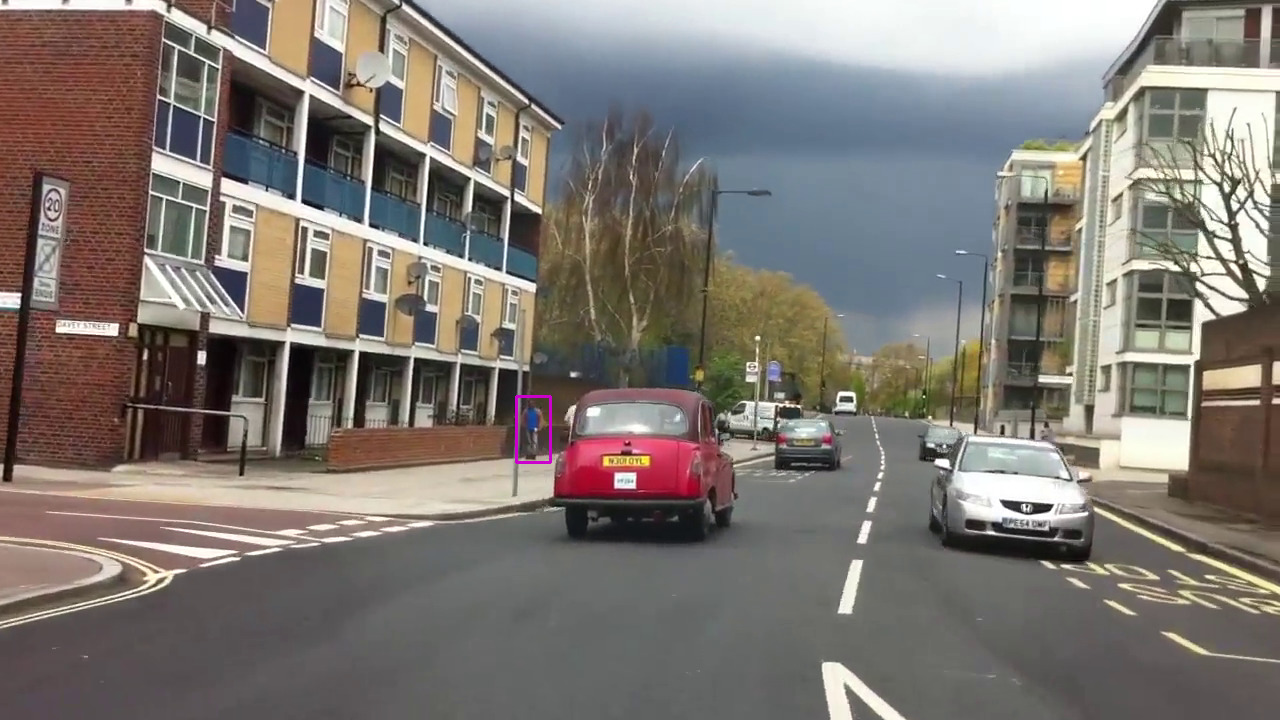} \\  \includegraphics[width=1.5in,height=1in]{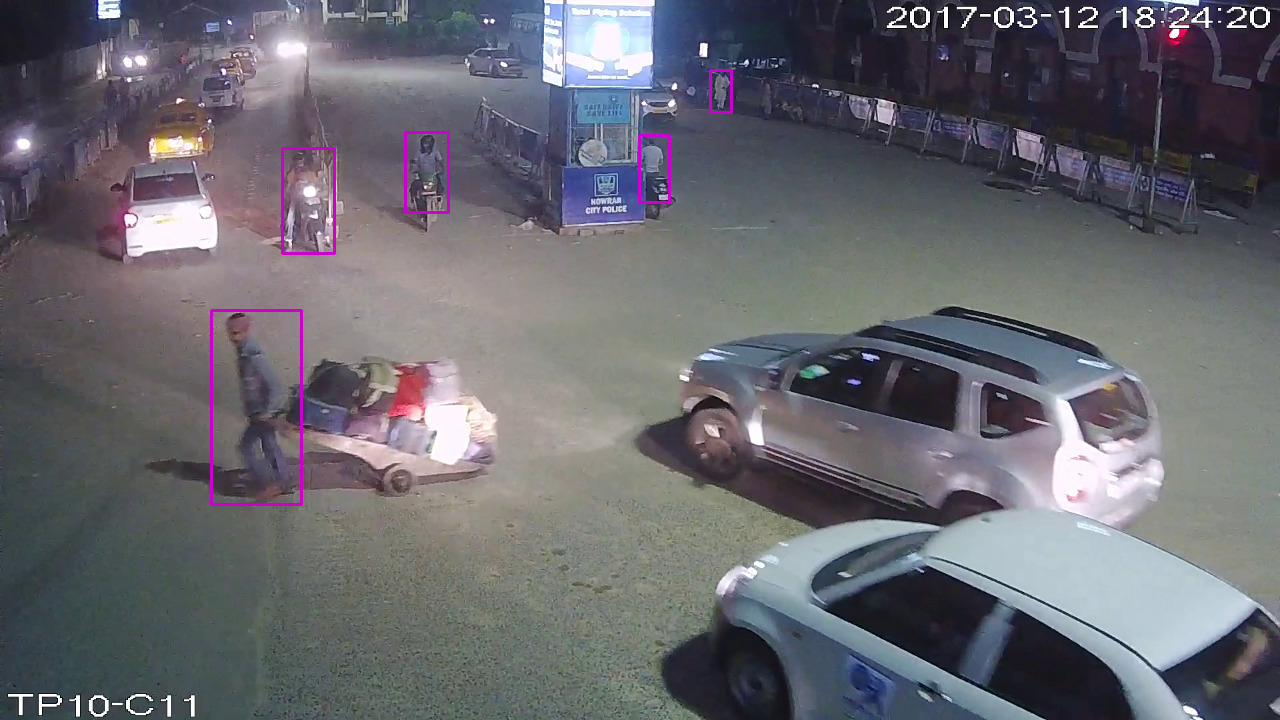} \hfill
     \includegraphics[width=1.5in,height=1in]{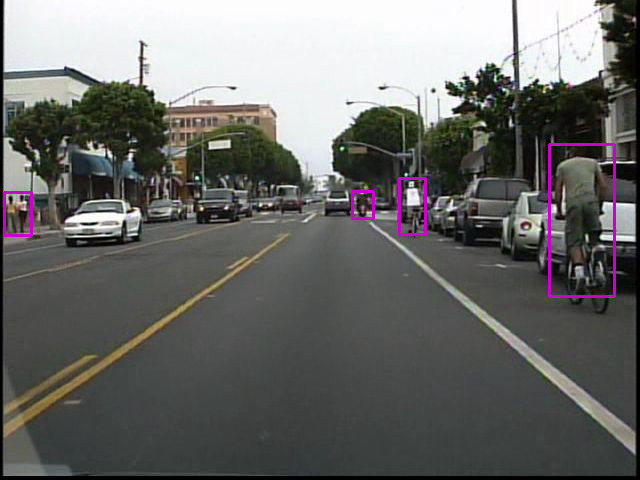} \\
    \includegraphics[width=1.5in,height=1in]{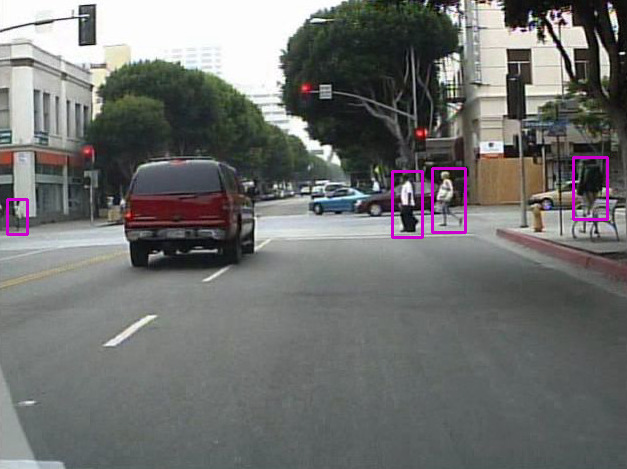} \hfill
    \includegraphics[width=1.5in,height=1in]{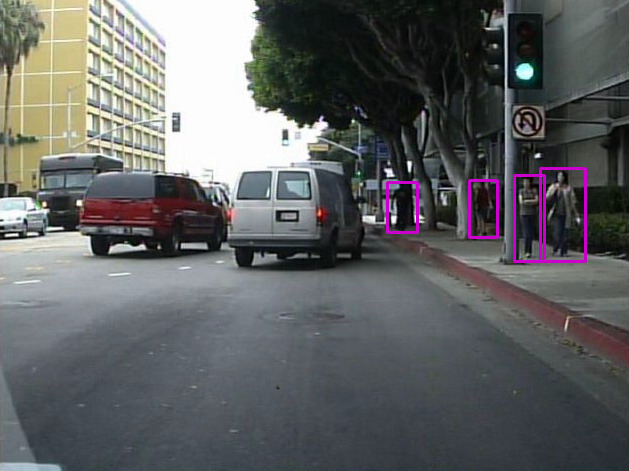}
    \label{fig:outputs}
    \vspace{-0.10in}
    \caption{Some Output Image Generated from SaYwF.}

\end{figure}

\vspace{-0.14in}

 The outcome of our experimentation (shown in Table \ref{tab:exp_result}.) clearly suggests that the proposed model SaYwF obtained better accuracy. Also, its processing speed (FPS) is very close to the DNC benchmark. On the other hand, SaYwF did not perform up to the mark in terms of FPS for the newly created dataset as the video resolution of the new dataset is nearly double to that of CPD. We plan to do studies to improve the performance of our model for high resolution video processing. 

As FPS depends on the stride of the sliding window we also performed several experiments to observe the variation of FPS against stride and also variation of FPS against MR to justify the FPS MR trade-off. This is illustrated in Fig. \ref{fig:stride_plot} where MR and FPS are normalized in range (0,1). Note that the distance of pedestrian from camera is categorized in three classes and they are \textbf{Near}, \textbf{Medium} and \textbf{Far}.



\begin{figure}[h]
\vspace{-0.17in}
\begin{tabular}{cc}
  \includegraphics[width=1.8in,height=1.7in]{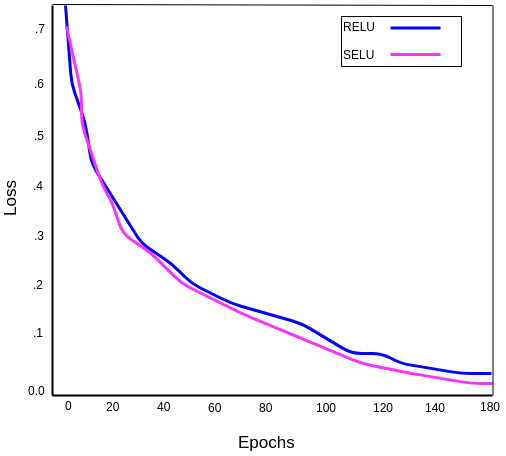}  \\
\end{tabular}
\vspace{-.19in}
\caption{Comparison of losses provided by RELU vs. SELU.}
\label{fig:selu_relu}
\vspace{-0.15in}
\end{figure}

As mentioned earlier, we have opted for \textbf{SELU} activation function over \textbf{RELU}, a more popular variety. This choice has been prompted by extensive experimentation and here, in Fig. \ref{fig:selu_relu}, we have shown the comparative performance in terms of network loss on the combined training sets of the three databases used in the present study.


Another set of experiments has been performed with different train set and test set combinations. We have used negative mining to increase the volume of the train set of previously specified binary classifiers. Here, we have taken false negative from SaYwF output and included them as negative samples in the training set to improve overall accuracy. These results are provided in Tables \ref{tab:exp_result_2} and \ref{tab:exp_result}. Also, the recall values of both the classifiers are provided in Table \ref{tab:recall_stage}.

\begin{table}[h]
    \vspace{-0.10in}
    \caption{MR of different models corresponding to different combinations of train and test sets}
    \vspace{-0.12in}
    \begin{tabular}{|c|c|c|c|c|}
    \hline
    \textbf{Model} & \textbf{Test $\downarrow$} & \multicolumn{3}{c|}{\textbf{Train}}\\
    \cline{3-5}
     &  & CPD & CityPersons & ISIPD\\ \cline{2-5}
     \multirow{3}{*}{DNC} & CPD & 39.37 & 41.77 & 35.53 \\ \cline{2-5}
     & CityPersons & 43.19 & 44.92 & 33.79  \\ \cline{2-5}
    & ISIPD & 45.48 & 48.67 & 33.43 \\ \cline{2-5}
    \hline
    \multirow{3}{*}{googleNet} & CPD & 37.49 & 38.93 & 35.53  \\ \cline{2-5}
    & CityPersons & 40.93 & 42.63 & 31.08  \\ \cline{2-5}
    & ISIPD & 43.51 & 46.18 & 31.58 \\ \cline{2-5}
    \hline
    \multirow{3}{*}{SaYwF(Our)}& CPD & 37.28 & 38.43 & 32.31  \\ \cline{2-5}
    & CityPersons & 39.58 & 40.52 & 29.31  \\ \cline{2-5}
    & ISIPD & 42.98 & 44.73 & 29.93  \\ \cline{2-5}
    \hline
    \end{tabular}
    \label{tab:exp_result_2}
    \vspace{-0.18in}
\end{table}

  \begin{table}[!h]
  \vspace{-0.10in}
   \caption{Experimental results (MR and FPS) of different models trained using CPD, CityPersons, ISIPD datasets}
   \vspace{-0.12in}
    \begin{tabular}{|c|c|c|c|c|c|c|}
    \hline
    \multirow{3}{*}{\textbf{Architecture}} & \multicolumn{6}{c|}{\textbf{Dataset}}\\
    \cline{2-7}
     & \multicolumn{2}{c|}{CPD} & \multicolumn{2}{c|}{City Person} & \multicolumn{2}{c|} {ISIPD} \\ \cline{2-7}
    & MR & FPS & MR & FPS & MR & FPS \\ \hline
    DNC & 25.39 & 15 & 28.77 & 7 & 31.43 & 10\\ \hline
    googleNet & 21.49 & 21 & 25.93 & 10 & 28.58 & 13\\ \hline
    SaYwF & 18.11 & 20 & 23.78 & 8 & 27.33 & 11\\ \hline
    \end{tabular}
    \label{tab:exp_result}
\end{table}

\begin{table}[!h]
\vspace{-0.21in}
\caption{Recall at different stages of classification.}
    \vspace{-0.12in}
    \begin{tabular}{ |c|c|c| }
    \hline
    Phase & Pedestrian Recall \\
    \hline
    \hline
    $C_z$ (Phase I) & 93.04 \\
    \hline
    $C_p$ (Phase II) and NMS (Phase III) & 88.01 \\
    \hline
    \end{tabular}
    \label{tab:recall_stage}
\end{table}

\begin{figure}[!h]
\begin{tabular}{cc}
   \includegraphics[width=1.6in,height=1.5in]{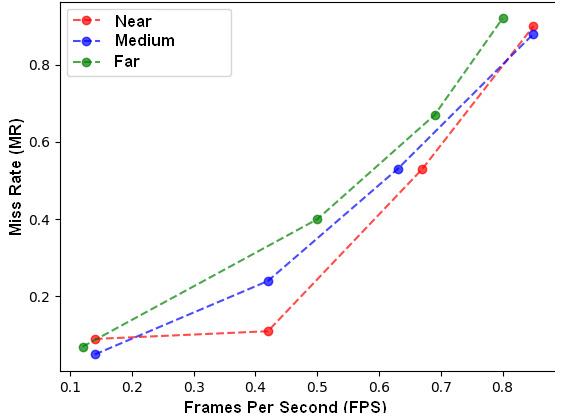}
   \includegraphics[width=1.6in,height=1.5in]{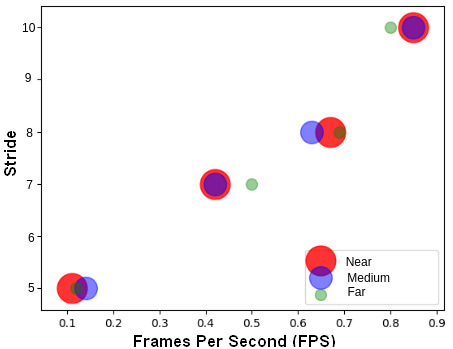}
\end{tabular}
\vspace{-.2in}
\caption{Dependency between MR, FPS and stride size.}
\label{fig:stride_plot}
\end{figure}


\vspace*{-.20in}

Here it may be noted that the first stage eliminates a large number of candidate windows (non-potential zones) for improving the processing time. In this phase one can also use some non-CNN based algorithms like exhaustive search \cite{harzallah2009combining}, selective search \cite{uijlings2013selective} or BPM \cite{felzenszwalb2010object} techniques, but our experiment shows that $C_z$ classifier based approach is both accurate and faster. In the first stage (that uses the $C_z$ network) it takes 7.2 ms to detect possible potential zones among 16 regions of an input frame (refer to Sec \ref{sec:phaseI}). The second stage (with $C_p$ classifier) takes around 40.7 ms to predict all bounding boxes. Overall execution time is 52 ms (including NMS) for a 640 $\times$ 480 frame consequently yielding 20 FPS processing time.

\section{Conclusions}\label{sec:conclusion}

Considering real life applications of pedestrian detection systems, its execution in real time with high accuracy is crucial. However, there is hardly any available system that is optimized with respect to both execution speed (FPS) and detection accuracy (measured in terms of miss rate or MR). In view of the same we have designed the proposed Seek and You will Find (SaYwF) architecture. It first looks (Seek) for potential zones of a video frame where a pedestrian or its part may exist. Next, it identifies (You will Find) the part of such a zone where the pedestrian or its part actually appears. An instance of a pedestrian may lie in one or more such neighbouring regions. The final non-maximum suppression stage combines the parts of these regions to obtain the bounding box of the instance of a pedestrian. For effective training of the proposed model we have developed a new sample dataset (ISIPD) to capture wide variations among the training samples. This database will be distributed freely for academic research purposes.


\section*{Acknowledgement}We gratefully acknowledge the support of NVIDIA Corporation with the donation of the Titan Xp GPU used for this research. 

\bibliographystyle{ACM-Reference-Format}
\bibliography{acmart}

\end{document}